# The Limits of Learning from Pictures and Text: Vision-Language Models and Embodied Scene Understanding

Gillian Rosenberg[1], Skylar Stadhard[1], Bruce C. Hansen[2], Michelle R. Greene[1]

[1] Barnard College, Columbia University  [2] Colgate University

## Abstract

What information is sufficient to learn the full richness of human scene understanding? The distributional hypothesis holds that the statistical co-occurrence of language and images captures the conceptual knowledge underlying visual cognition. Vision-language models (VLMs) are trained on massive paired text-image corpora but lack embodied experience, making them an ideal test of the *distributional hypothesis*. We report two experiments comparing descriptions generated by 18 VLMs to those of over 2000 human observers across 15 high-level scene understanding tasks, spanning general knowledge, affordances, sensory experiences, affective responses, and future prediction. Because many tasks lack ground truth answers, we developed a Human-Calibrated Cosine Distance (HCD) metric that measures VLM output similarity to the distribution of human responses, scaled by within-human variability. In Experiment 1, VLMs approached human-level performance on general knowledge tasks, but showed a robust deficit for affordance tasks that resisted prompt engineering and did not improve with newer model releases. In Experiment 2, we tested six mechanistic hypotheses for explaining this affordance gap, finding that the deficit was structural rather than stylistic and was not resolved by providing explicit spatial information. Corpus analyses revealed that image captioning datasets contain sparse agent-addressed affordance language, consistent with Gricean accounts of why embodied knowledge may be systematically underrepresented in language. Together, these findings suggest that distributional learning from images and text is insufficient for affordance-based scene understanding, implying that some dimensions of human visual cognition may require the kind of agent-centered, three-dimensional experience that no photograph or caption can encode.

# Introduction

Consider the photograph in the left-hand column of Figure 1. Immediately and effortlessly, you can name its general category (a thrift store), and name some of the objects in it (the American flag, or vinyl records, for example). Although categorization and object identification are considered "core" components of human (DiCarlo, Zoccolan, & Rust, 2012) and machine (Russakovsky et al., 2015) visual understanding, human scene understanding is characterized by a deep inferential chain of thought that allows us to reason about what actions could be taken in the scene, what it would feel like to be in the scene, predict what will happen next, and speculate about what happened before. Where did this knowledge come from? What information is sufficient to develop these abilities? Recently-developed vision-language models (VLMs) provide a method for testing these questions. VLMs are trained with a massive amount of language content and photographs, but they are not embodied, nor do they have the capacity to explore the world to discover its causal connections. Therefore, if a VLM is able to succeed in high-level scene understanding tasks, we can infer that those tasks can be learned from the statistical structure of language and images. The answers to these questions serve both a basic question about human cognition (what are the necessary inputs for scene understanding?), as well as a practical one (can VLMs serve as models of human scene perception?). On the practical side, VLMs are already being used in psychological sciences to scale cognitive research by annotating items such as ground truth categories (Rong, Gifford, Düzel, & Cichy, 2025), emotional features like facial expressions (Niu, El-Tawil, Romana, & Provost, 2025), and semantic descriptions (Murlidaran & Eckstein, 2025). As use of VLMs as a modelling tool increases, it is necessary to assess the quality of this model generated content.

The *distributional* hypothesis posits that language captures an impressive amount of human conceptual knowledge, and predicts that much of high-level scene knowledge should be recoverable from the statistical structure of language and images alone. A strong version of this hypothesis would predict high performance across a range of scene understanding tasks. There are reasons to take the distributional hypothesis seriously. For example, blind individuals share color category structure with sighted individuals, despite never visually experiencing color (Kim, Aheimer, Montané Manrara, & Bedny, 2021), and VLMs share similar performance patterns on sensory concepts with and without visual training (Marjieh, Sucholutsky, van Rijn, Jacoby, & Griffiths, 2024), suggesting that at least some visual concepts can be learned without visual experience. Additionally, even higher-level cognitive abilities, such as theory of mind, seem to be largely captured by a massive amount of language training (Trott, Jones, Chang, Michaelov, & Bergen, 2023). But there is also reason for skepticism of the distributional hypothesis — many large language models (LLMs) improve with visual concept grounding (Xu et al., 2025), showing that sensory experience augments the information provided by language, and suggests that other grounding may lead to further improvements. Further, VLMs demonstrate persistent deficits in some basic visual tasks (Tangtartharakul & Storrs, 2026; Tong et al., 2024), so it may be the case that even massive-scale learning of two-dimensional visual inputs can never fully generalize to the three dimensional world. A second possible issue is in the pragmatics of human communication. Humans tend not to express ideas that are self-evident to other humans (Grice, 1991). Therefore, if LLMs and VLMs are trained on human communication patterns that systematically compress information that is obvious to embodied agents, no amount of text-image training would allow a VLM to learn about certain tasks. Finally, empirical evidence shows that both typically-developing children and children who have had their sight surgically restored use self-motion to parse and segment objects (Ostrovsky, Meyers, Ganesh, Mathur, & Sinha, 2009; Soska, Adolph, & Johnson, 2010), suggesting that humans may require three-dimensional experience to solve the problem of object segmentation.

In this work, we tested the distributional hypothesis across a broad range of high-level scene understanding tasks. These included core, general knowledge tasks, such as categorization and general scene descriptions, as well as affordances, sensory experiences, affective experiences, and future forecasting. Critically, these task classes differ in the degree to which they depend on embodied or agent-centered knowledge, yet human observers have been shown to perform these tasks with scenes quickly and easily. Thus, this work differentiates itself from traditional benchmarking in two ways: first by examining patterns of performance across tasks to determine what makes a task easy versus hard for VLMs, and second, by selecting tasks that go beyond information that is grounded in objects, attributes, and relations and towards those that require predictive, counterfactual, and inferential knowledge. A natural consequence of using these types of tasks is the need to define performance in a human-centered way. Because some tasks have no ground truth, we cannot evaluate VLMs using standard accuracy metrics. Instead, we assess whether VLM outputs fall within the distribution of human responses, suggesting that the models reason about scenes in a human-like way.

This paper is divided into two experiments. In Experiment 1, we compare descriptions from 18 VLMs to human descriptions across 15 high-level scene understanding tasks, arranged hierarchically in five task types. We developed a method for establishing whether a VL model response is within the human noise ceiling. Therefore, VLM success is defined within quantitative similarity to the human distribution of responses. We examined how robust the task-specific patterns are to changes in factors such as prompt, model family, and embedding type. In Experiment 2, we narrow our investigation to explore a key finding from Experiment 1 that there is a persistent gap between VLM descriptions and human descriptions for affordance tasks. We evaluate six hypotheses for why these differences persist for these specific sets of tasks. We find that VLMs approach human-level performance on general knowledge tasks but show a persistent deficit for affordance tasks that resists prompt engineering, does not improve with newer models, and reflects structural rather than superficial differences.

# Experiment 1: Human and VLM Scene Descriptions

The goal of Experiment 1 was to assess whether VLMs demonstrate human-like visual scene understanding across a range of high-level tasks. If VLMs are successful at a task, then the task is learnable from the correspondence of distributional semantics and two-dimensional photographs. Many high-level tasks require counterfactual and causal reasoning, or invoke subjective qualitative experience, and thus do not have ground truth answers. To quantify VLM similarity to the distribution of human descriptions, we created a human-calibrated cosine distance metric. Throughout, we assessed the robustness of task related differences across a number of factors such as prompt type, model type, and embedding type. Our central finding of this experiment is that affordance tasks consistently had the least similarity between VLM and human descriptions, even across prompt types and time.

Our first step was to obtain human and VL model scene descriptions across a diverse set of scene understanding tasks and to quantitatively compare their similarities and differences.

## Methods

### Image Selection

We selected a set of 40 images of real-world scenes spanning indoor, outdoor, natural, urban, public, and private spaces. This sample size balances the need for sufficient statistical power with the practical constraints of collecting detailed descriptions across 15 different tasks, 18 different VLMs, 10 human observers, and three prompting conditions, for a total of over 3.2 million different descriptions. Photographs were selected from an internal lab dataset that has not yet been released to the public. As such, these photographs constitute new content not present in the training corpora used by any model. This allows us to conduct a stronger test of how well models can generalize to new content. Images depicted quotidian real-world settings that do not reflect prototypical scene categories, limiting the extent to which models could rely on learned category-text associations.

### Tasks

Our goal was to create a battery of high-level scene understanding tasks that range from objective and knowledge-based to predictive and embodied, capturing the highly inferential and multi-modal nature of scene understanding. Our battery consisted of 15 tasks that were divided into five groups: general knowledge, affordances, sensory experiences, affective experiences, and future casting. We will introduce each group in turn.

The general knowledge group represents objective facts about the scene's category, objects, and basic layout. These tasks were chosen to reflect what some have termed "core" human visual cognition (DiCarlo et al., 2012). The specific tasks within this group included naming the scene's basic-level category (Tversky & Hemenway, 1983), naming its major objects (Friedman, 1979), and providing a general scene description (Fei-Fei, 2007). As computer vision models are often optimized for these tasks (Krishna et al., 2017; Russakovsky et al., 2015), we expect that VLMs should approach human-level performance in this group.

Affordances, the set of possible actions that an agent could take in a scene, are foundational to human visual understanding (Gibson, 1986). These tasks were selected because scene affordances are understood by human observers very quickly and easily (Greene & Hansen, 2025; Greene & Oliva, 2009), despite requiring physical counterfactual reasoning. The specific tasks in this group included listing a general list of possible actions one could take in the scene (Greene, Baldassano, Esteva, Beck, & Fei-Fei, 2016), providing a description of possible navigational paths (Bonner & Epstein, 2017), and describing where a human could sit in each scene. For VLMs to succeed in these tasks, it will be necessary to reason about a body-scaled agent in a 3D scene from combining knowledge from 2D picture representations and distributional language training.

The sensory experiences group probes knowledge that is grounded in cross-modal perception rather than visual input alone. Human sensory perception is highly multimodal, and humans are able to reason about what they would hear, smell, and touch using only visual input (Fassnidge & Freeman, 2018; Gottfried & Dolan, 2003; Keysers et al., 2004). The specific tasks in this group included generating a list of sounds, smells, and touch experiences they would experience in each scene (Fassnidge & Freeman, 2018; Gottfried & Dolan, 2003; Keysers et al., 2004), and providing an assessment of how loud each scene would be if one were there (@ Fassnidge & Freeman, 2018), and estimating of the physical temperature of each environment (Greene & Oliva, 2009). VLMs may not perform well in these tasks, given that they lack access to cross-modal grounding.

Affective responses to scenes are ecologically fundamental, shaping approach and avoidance behavior in ways that are tightly coupled to scene identity (Russell & Pratt, 1980). Tasks in the affective experiences group included describing emotional experiences people would have while in the environment (Olofsson, Nordin, Sequeira, & Polich, 2008), analyzing the extent to which they would feel safe in the environment (Schroeder & Anderson, 1984), and assessing how aesthetically pleasing they found the environment (Vessel & Rubin, 2010). Because affective responses to environments are grounded in embodied experience of being *in* a space, they may represent dimensions of scene knowledge not captured by vision-language co-occurrence learning.

Finally, scenes provide rich contextual information that supports prediction of future states (Bar, 2004; Freyd, 1987). Consequently, the tasks in the future casting group required participants to mentally simulate different possible futures for each scene, considering how quickly the scene's appearance is changing (Greene & Oliva, 2009), the amount of mystery versus predictability in the scene (Kaplan & Kaplan, 1989), and to describe how the scene will appear five minutes into the future (Fouhey & Zitnick, 2014). Each of these tasks requires temporal extrapolation beyond the current window, requiring causal knowledge of how environments and agents will interact. The precise instructions provided to human participants for each task are found in Table 1.

Table 1: Instructions given to participants for each of the 15 tasks. We classified the 15 tasks into five categories: general knowledge, affordances, sensory experiences, affect, and future-casting.

| Task Group | Task | Human Prompt |
|---|---|---|
| General knowledge | General description | Please describe each photograph in a way that would allow someone to identify the picture solely from the description. For example, 'office' would not be enough detail to tell what kind of an office is in the photograph, but a more extended description such as 'View of a messy desk in a home office. There is a laptop computer with two computer monitors. A framed photograph of a family is located to the right side of the monitors' would enable someone to identify the unique features of the given photograph. |
| General knowledge | Basic-level category | Please describe in one word the general type of scene that is captured by the photograph. Examples include: 'forest,' 'street,' 'office,' 'beach'... |
| General knowledge | Objects | List 10 things you see in the scene. If there are more than 10, list the 10 most important things. If there are fewer than 10, list all the things. Examples of things include: 'sky,' 'cup,' 'book,' 'water'... |
| Affordances | General affordances | Please write a list of all the things that one might do in this scene using verbs. Examples of verbs include: 'walk', 'sleep', 'eat', 'draw a picture,' 'go sailing'... |
| Affordances | Navigation | Imagine that you are telling a blind person how to walk through this scene. Write a short (1-3 sentence) description that could instruct such a person. |
| Affordances | Sitting | Imagine you are in the scene and you need to sit down to rest. Describe any and all surfaces where you might reasonably choose to sit and why. For example, "I might choose to sit on the couch in the back left corner of the image because it looks cozy, or I could sit on the straight back chair on the right side which is also available." |
| Sensory experiences | Multisensory | Imagine that you have entered the scene. Use your non-visual senses to describe the scene. What would you hear, what would you smell, what would feel, and, if applicable, what would you taste? |
| Sensory experiences | Loudness | How loud do you think it would be in the scene ranging from extremely quiet to extremely loud and why? In the quietest scene, you could hear a pin drop and, in the loudest scenes, you would have hearing damage if you didn't use protective equipment. Ex: "I imagine it would be pretty loud because it is a bustling street with cars, trucks, and people." |
| Sensory experiences | Physical temperature | Imagine you were in the scene. How hot or cold would it be? Try to avoid using specific numbers for temperature. Instead, say something more descriptive and include your reasoning for why you think it would be that hot or cold. Ex: it would be extremely hot because the sun is beating down on the concrete. |
| Affective | Emotions | Imagine yourself in the scene. Describe the emotions that you would feel and why. For example, when describing a serene, sunny beach scene you might 'I would feel joyful and relaxed because this reminds me of vacation.' |
| Affective | Safety | How safe would you feel walking alone in the scene pictured above and why? For example: I would not feel very safe because it is a very dark street and there does not seem to be anyone else or any cars around. |
| Affective | Aesthetics | How aesthetically appealing is the scene? Describe the elements of the scene that drove your answer. |
| Future casting | Transience | Some scenes are generally unchanging. Consider a janitor's closet: this scene would not appear differently until a janitor comes by to pick up some items, but this is rare. On the other hand, some scenes are constantly changing. Consider a busy airport terminal: this scene appears differently at every moment because people are always coming and going as flights board and deplane. How generally changing is this scene? |
| Future casting | Predictability | Scenes vary in their predictability. If you can only imagine one or two ways the scene might appear, then we would consider that a highly predictable scene. For example, a mountain range is highly predictable. It might appear differently if it's storming vs a sunny day, but for the most part, it always looks the same. However, if you could imagine the scene appearing many different ways, then that is highly unpredictable. For example, a subway station is highly unpredictable because it might be empty or crowded, a train might be arriving or not, the people in the station might be primarily dressed in business attire or sports fans in ball caps, etc. How predictable is this scene and why? |
| Future casting | Temporal | Imagine that you are witnessing the scene as it unfolds in real time. Please use your imagination to say how much the scene will change over the course of five minutes and what in the scene might appear differently as it changes. |

**Human descriptions**

A total of 2,043 English-speaking participants were recruited to provide scene descriptions from CloudResearch Connect (Hartman et al., 2023). 1,054 participants identified as women, 933 as men, 33 as non-binary, 6 as agender, 11 as members of non-listed gender minorities, and 6 as not disclosing their gender. Participants ranged in age from 18 to 83 years (mean=37.7 years, median=36 years, sd = 12.5). 1902 participants were located in the United States, 69 in the United Kingdom, 63 in Canada, and three each in Australia, Ireland, and New Zealand. Each task was launched as a separate experiment, and within each experiment, an individual could write as many descriptions as they wished that were still available from the pool. Each individual contributed between 1 and 41 descriptions (mean=2.9 descriptions, median=2 descriptions, sd = 3.03), and was compensated $0.25 per description. An additional 56 descriptions were collected but rejected and replaced because participants did not follow instructions or contained inappropriate language. Figure 1 shows representative descriptions for two example images across the 15 tasks.

**Vision-language models**

In total, we tested 18 different vision-language models. We aimed to get a range of model ages, as well as open- and closed-weight models in our analysis. Models were coded as open-weight if their trained weights were publicly downloadable for third-party local deployment; models without publicly downloadable weights (including API-only and enterprise-restricted releases) were coded as closed-weight. Models were from six families: GPT (o3, GPT-4o-mini, GPT-4-turbo, and GPT-5-mini); Claude (Haiku 3.0, Haiku 3.5, Sonnet 4.5, Opus 4.0); Gemini (Flash 2.0, Flash 2.5, Pro 2.5); Mistral (Mistral Medium, Magistral Medium, Pixtral Large); Llama (Llama 3.2 11B, Llava 1.6); and Deepseek (VL2-tiny, Janus Pro 1B). Technical specifications for all models can be found in Supplemental Item 1.

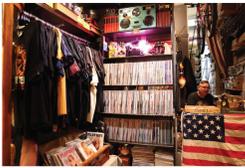
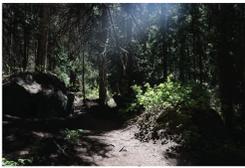

| | | | |
|---|---|---|---|
| **General** | General | A store with a big American flag on the shopkeep's desk. There are magazines and books available, along with some clothes | A walking trail in the middle of a forest. There is a giant stone boulder and a bunch of trees and greenery. |
| | Objects | person, flag, magazines, shirts, desk, shelves, radio, box, hangers, cameras | sky, tree, bushes, dirt, grass, red box, rocks, foliage, branch, leaves |
| | Categorization | store | forest |
| **Affordance** | Affordances | browse, shop, sift through, talk | walk, sit, hike, eat, run, jog, climb, draw, photograph |
| | Navigation | Walk forward a few steps. You may explore the merchandise on your left. There is a man behind a desk on your right. | You can walk forwards about 10 meters until you will reach a bunch of rocks. You then must move a few feet to the left to avoid the rocks. |
| | Sitting | I don't see many good surfaces to sit here. Maybe on the counter or on the floor, which isn't shown in the photo. | I would sit on one of the large rocks, so I'm not sitting directly on the ground. |
| **Sensory** | Multisensory | It would smell old and musty. It would feel a little warm and stuffy. | I hear birds chirping, and perhaps the wind is blowing through the trees. I would feel the wind on my skin and smell the dirt. |
| | Loudness | This would probably be fairly quiet because it looks like a thrift shop and those usually are pretty calm. | I imagine it would be pretty quiet because it looks to be in a forest with no one and nothing around. |
| | Temperature | I think it would be very warm because the sun is shining and it is a very crowded room | Warm because the sun is shining through the trees and plants all have leaves indicating it likely summer. |
| **Affective** | Emotions | I would feel out of place and annoyed that I feel out of place at this store that seems to be geared towards men. | I would feel as though I was in a fairy tale, I would feel lost in the moment, almost like in another world. |
| | Aesthetics | Not at all! Very crowded/visually noisy. I also don't appreciate the playboy magazines. | It's okay. The trees are beautiful. The patchiness of the grass sort of takes away from the surroundings. |
| | Safety | I would feel safe, although it looks cramped. It looks cozy and welcoming. It does seem a little clustered and claustrophobic but the colors and mix of decoration makes it feel homey | I would feel safe walking here but would be cautious as there could be wild animals or tripping hazards present. |
| **Future casting** | Transience | This scene is relatively generally changing, people will come and go to and from the shop. | I think it would be generally unchanging in the short term but long term it would change a lot with the seasons and weather. |
| | Temporal | A customer may likely appear to either browse/look at the items pictured. Or a customer may walk up to the worker sitting down to pay or ask a question. | This scene might change with the amount of sunlight shining, as the sun moves around the trees making it brighter or darker. |
| | Predictability | This scene looks fairly predictable. Small changes like clothing being moved, or magazines being moved can occur, but everything else seems static. | This scene is predictable. Hikers or animals may appear, but it will mostly remain unchanged |

Figure 1: Representative human descriptions for each of the 15 tasks for two example images.

**Prompts**

To ensure that no model or task was disadvantaged by a single prompting strategy, we tested each model under three prompt conditions. The first prompt set was identical to the prompt provided to the human observers (see Table 1). The second set of prompts was engineered within GPT-4o-mini to produce the most qualitatively human-like responses, as assessed qualitatively by the authors (GR and SS). The third prompt set was created by asking each model to produce a prompt that would elicit the most human-like responses on each task. Supplemental Item 2 contains the meta-prompts used to elicit prompts from all models. The final prompts provided by each model are in Supplemental Item 3. For each model and prompt, we elicited 100 unique descriptions for each of the 40 images, using the default temperature setting for each model.

**Embeddings**

Each description (human and VL model) was embedded using three embedding models: MPNet (Song, Tan, Qin, Lu, & Liu, 2020), CLIP (Radford et al., 2021), and RoBERTa (Liu et al., 2019) to ensure that our results were not an artifact of any single embedding space, but instead replicated across complementary representational biases (sentence-level semantic similarity in MPNet/RoBERTa versus vision–language alignment in CLIP). Specifically, MPNet excels at sentence-level semantic similarity, while RoBERTa captures contextual language representations. Patterns that persist over all embedders likely reflect differences in VL model knowledge gained from their pretraining.

**Human-calibrated cosine distance (HCD)**

Many of the tasks we consider here require the observer to make predictions or write about subjective experiences for which there is no ground truth. Our goal, therefore, is to establish the extent to which VLMs reason about visual scenes in a human-like manner. To this end, we created a human-calibrated cosine distance (HCD) between human and VLM embedding vectors. This metric measures distance relative to within-human variability, and ranges from statistical replaceability (i.e., as similar to human descriptions as humans are to each other) to being more dissimilar than a human description of an entirely different image.

The HCD is computed as follows:

$$HCD = \frac{d_{HM} - LB}{UB - LB}$$

Where $d_{HM}$ is the median cosine distance between human and model embeddings, $LB$ is the lower bound of human-human variability, and $UB$ is the upper bound. The human-model distances ($d_{HM}$) were computed as cosine distances between L2-normalized embedding vectors from human and VL model descriptions for each of the 15 prompt tasks (computed separately for each image and embedding type). We established upper and lower bounds relative to human-to-human description distances. First, we computed a per-image human "center" as the unit-normalized mean of the human embeddings. We then computed a lower bound to estimate within-image human–human variability: for each image, we iteratively left out one human observer and computed the cosine distance between the held-out human embedding and the centroid of the remaining human embeddings (leave-one-out, LOO). If VL–human distances approach this lower bound, VL model descriptions can be considered statistically indistinguishable from human responses. We also computed an upper bound intended to estimate across-image dissimilarity: for each image, we computed cosine distances between its human embeddings and the centroids of all other images, and defined the upper bound as the 95th percentile of these cross-image distances across images. If VL–human distances approach this upper bound, the VL descriptions are as dissimilar to the target image as a human description would be to an entirely different image.

A human-calibrated cosine distance (HCD) of zero indicates full human replaceability (i.e., a model description that is statistically indistinguishable from human descriptions), while a value of one indicates a description that is as different from the given image as an entirely different image. Furthermore, negative values can indicate that model descriptions cluster more tightly around the human centroid than individual human responses do, reflecting response homogeneity, and values greater than one indicate catastrophic failures that exceed the "different image" benchmark.

**Natural Language Processing (NLP) Metrics**

We assessed general aspects of human and VL model descriptions using more traditional NLP metrics as well. Using a pipeline that previously assessed the complexity of scene descriptions (Aronson, Adkins, & Greene, 2025; Greene & Trivedi, 2023), we extracted three features from each description: number of words, lexical entropy, and the average cosine similarity between word pairs in each description, as evaluated by Word2Vec (Mikolov, Sutskever, Chen, Corrado, & Dean, 2013). Finally, to quantify subjective information (e.g., emotions or attitudes), we performed a sentiment analysis to obtain an average sentiment score for each image within a task and within a model using the NLTK rule-based sentiment analysis tool, Vader (Hutto & Gilbert, 2014).

## Results

**Human Descriptions Are More Efficient**

We found that VL descriptions are longer but more lexically redundant than human descriptions, whereas humans achieve greater lexical diversity at matched length. On average, VL model descriptions were longer, with a human-generated response being 14.4 words long while an average VL-generated response was 56.3 words long (see Figure 2A). We modeled log-transformed description length using a linear mixed-effects model with random intercepts for image and task. This model confirmed that VLM descriptions were significantly longer than human descriptions ($\beta = 1.10$ log-words, SE = 0.038, $t(33550) = 29.21$, $p < 2e-16$).

Although VL models produced much more text overall, we found that the lexical variability of this text was lower than that of human descriptions. We fit a linear mixed effects model predicting lexical entropy from description type (humans versus VL). Because entropy grows with description length, we included centered log description length as a covariate, along with an interaction term for description type. As before, we included random intercepts for image and task. As expected, we found that entropy grows with description length ($\beta = 0.686$, $p < 2e-16$), and that at the mean length, VL model descriptions had lower entropy than human descriptions (VL: $\beta = -0.500$, $p < 2e-16$, see Figure 2B). Finally, VL model entropy increased more steeply with length than human entropy (Type x Length interaction: $\beta = 0.190$, $p = 1e-8$). In practice, over the range of observed description lengths, humans remain more lexically diverse than VL model outputs at matched length. This pattern is further supported by humans' higher type-to-token ratio compared to VL models (M=0.89 versus 0.78, $\beta = -0.11$, $p < 2e-16$, see Figure 2C). Finally, we examined the average cosine similarity between pairs of concepts within each description, as represented with Word2Vec (Mikolov et al., 2013). Overall, we found no main effect of description type (human versus VL, $\beta = 0.0036$, $p = 0.204$, see Figure 2D), but we found a significant effect of description length ($\beta = -0.025$, $p < 2e-16$) as well as an interaction between Type and Length ($\beta = 0.009$, p=0.003), suggesting that Word2Vec cosine similarity decreased with description length, indicating increasing semantic diversity, but this decrease was significantly weaker for VL model text than for humans, suggesting that additional VL words add proportionally less new conceptual content.

Sentiment analysis revealed that the overall tone of VLM descriptions was much more positive compared to human descriptions. The sentiment analysis score for each image, prompt, and task ranged from -100 (very negative) to +100 (very positive). We fit a linear mixed effects model predicting sentiment from description type (humans versus

VL), and found that VL-generated texts scored much more positively than human-generated content (human: 9.67, VL: 26.36, $\beta$ = 16.69, p < 2e-16, see Figure 2E).

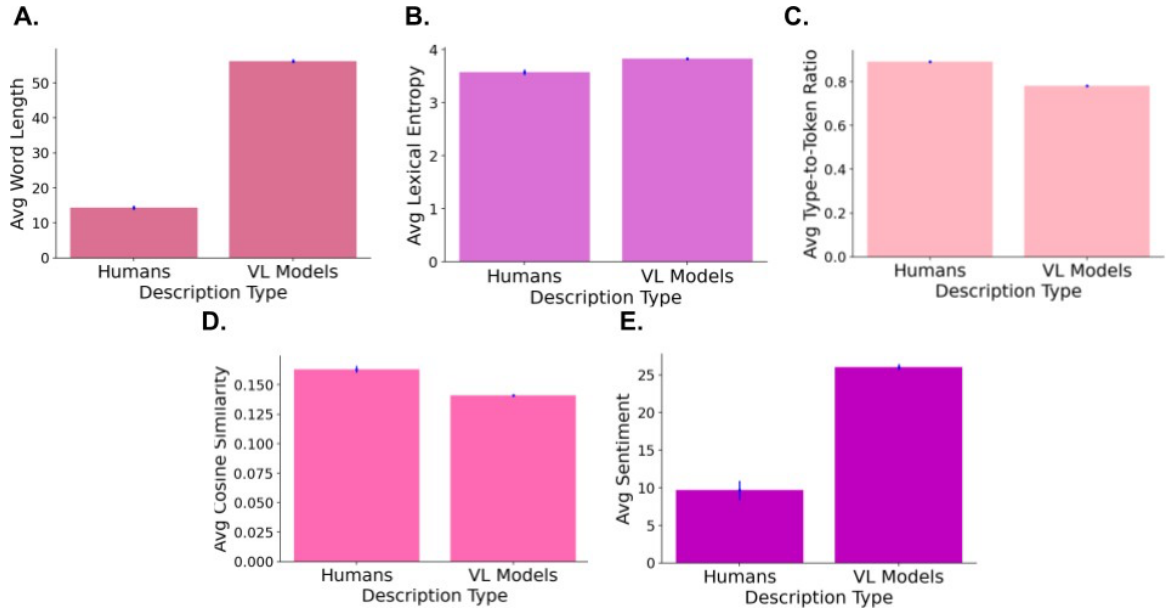

Figure 2: Comparisons between humans and VL models on NLP metrics taken from their respective generated descriptions. (A) Average description length was longer for VL models than humans. (B) Average Lexical Entropy was smaller for VL models than humans. (C) Average Type-to-Token Ratio was larger for humans than VL models. (D) Average Cosine Similarity was larger for humans than VL models until equating for description length (see text). (E) Average sentiment was more positive for VL models than humans. Error bars represent 95% confidence intervals.

**Human Descriptions are Sensitive to Task Hierarchies**

The representational geometry of human-generated description embeddings clustered hierarchically according to both specific task and task group (see Figure 3, left), rather than individual image identity (Figure 3, right). We quantified the apparent clustering structure using a k-nearest-neighbor (kNN) purity metric on the original embedding cosine distances at both the fine (15 tasks) and coarse (five superordinate groups) label levels. We first reduced the embedding dimensionality to the first 100 principal components for stability. Then, for each point $i$, we identified its k nearest neighbors (excluding itself) and calculated the fraction sharing $i$'s label, averaging across points for a final purity metric. We set k to equal 10% of the class size for each label (and have verified the robustness of this metric by sweeping k between 10% and 50%). We found that human scene descriptions had a kNN purity of 80% for the fine-grained tasks, and 91% for the superordinate task groups.

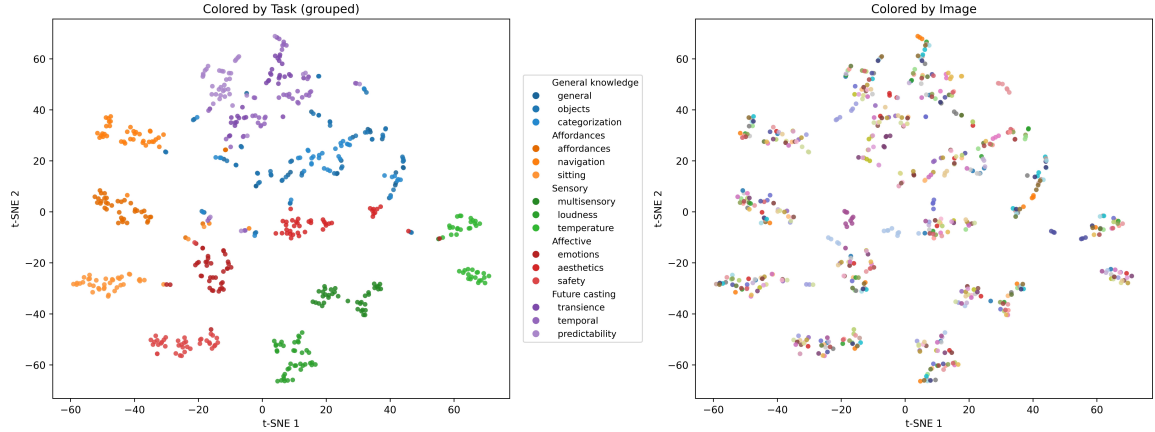

Figure 3: t-distributed stochastic neighbor embedding (t-SNE) projections of embeddings (MPNet) colored by task (left) and image (right). Embeddings cluster by task as opposed to image.

We conducted the same analysis for our 18 models (selecting the best-performing model from each VL family for visualization in Figure 4). As shown in Figure 4, we observed a different pattern of task clustering for VL models compared to humans: whereas fine-grained purity levels were similar (80% for both), the model embeddings showed little additional structure at the coarse level: whereas human embeddings became markedly more homogeneous when labels were collapsed into the five superordinate groups (82% for models versus 91% for humans). To directly test whether humans and models diverged more at the coarse than fine level, we computed a delta score for each model, $\Delta = (H - M)_{coarse} - (H - M)_{fine}$, where positive values indicate greater human–model divergence at the coarse level. All 18 models yielded positive delta values (range: 0.020–0.114), and the distribution of deltas was significantly greater than zero (Wilcoxon signed-rank test: V=171, p=7.63e-6). This pattern suggests that, in contrast to humans, VL models do not generate scene descriptions in a way that reflects the task hierarchy, yielding weaker within-task cohesion and greater mixing across higher-level groupings.

**Human-Calibrated Cosine Distances Show that VLMs Struggle with Affordance Tasks**

Across all task groups, we found that affordance tasks were uniquely difficult for VLMs. We examined a fully factorial design comprising HCDs across 18 models x 15 tasks x 3 prompt structures x 3 encoding models, yielding 2,430 observations. Figure 5 shows histograms of human-calibrated distances (HCDs), along with kernel densities, for task group, task generality, model family, and prompting strategy. Figure 6 shows representative descriptions for two example images across a spectrum of HCD values.

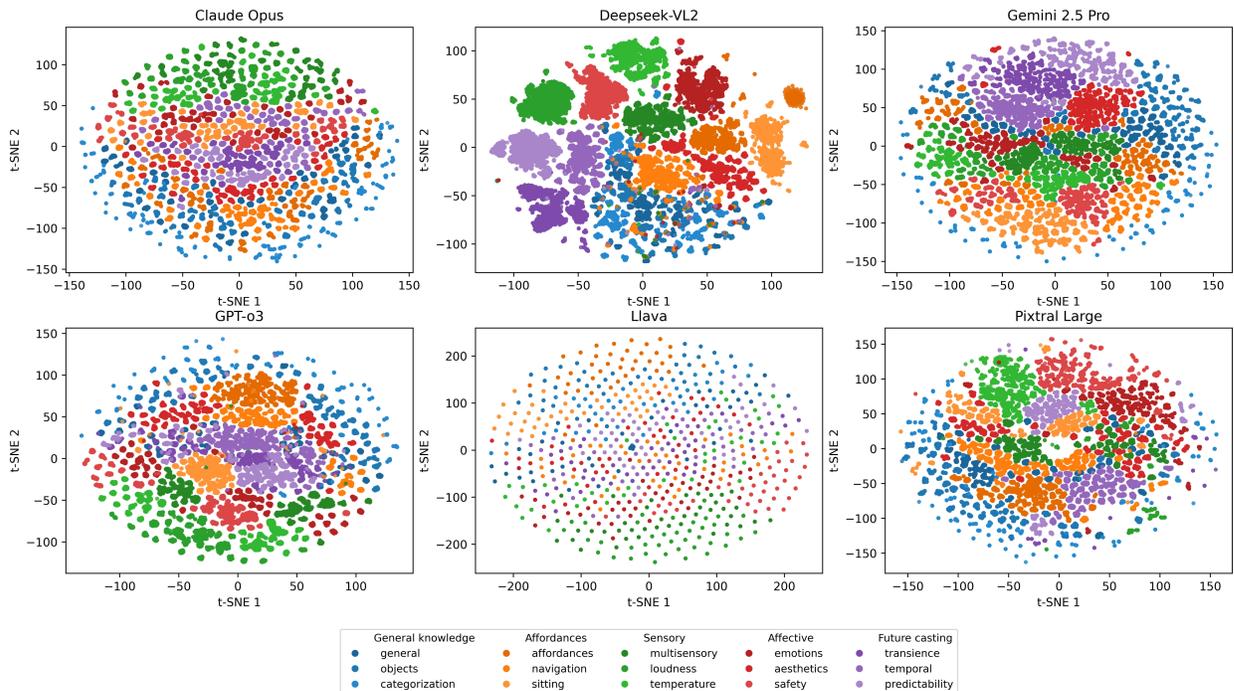

Figure 4: t-SNE projections of MPNet embeddings, colored by task and task group. One model from each family (top performing, according to human-calibrated cosine distance) is shown for clarity.

To assess task difficulty and prompting effects, we fit a crossed mixed model with fixed effects for Prompt type, task structure (Task Group and General-vs-Specific status), their interactions, and included random intercepts for Model and Task to account for baseline differences across the 18 vision–language models and 15 tasks.

Under the baseline (human) prompt, HCD was higher for affordances ($\beta$ = 0.171, 95% CI [0.04, 0.30], t(13.35) = 2.59, p = 0.022) and sensory tasks ($\beta$ = 0.163, 95% CI [0.03, 0.29], t(13.35) = 2.47, p=0.028) relative to general knowledge, whereas affective and future-casting tasks did not differ reliably from baseline. Affordance queries were the most difficult overall: human-calibrated distance was higher for affordances than the mean of the other four task categories (planned contrast: $\Delta$ = 0.178, SE = 0.047, t(9) = 3.76, p = 0.0045). This corresponds to affordance descriptions being ~18 percentage points farther from the human lower bound (as a fraction of the human range) than other task categories. Task generality did not significantly predict HCD ($\beta$ = 0.02, 95% CI [-0.07, 0.11], t(13.35) = 0.49, p = 0.63), suggesting that VLMs do not struggle to produce more specific scene information.

The model's intercept, corresponding to the human prompt, the general knowledge task group, the general tasks (versus specific) was at 0.18 (95% CI [0.05, 0.30], t(20.71)

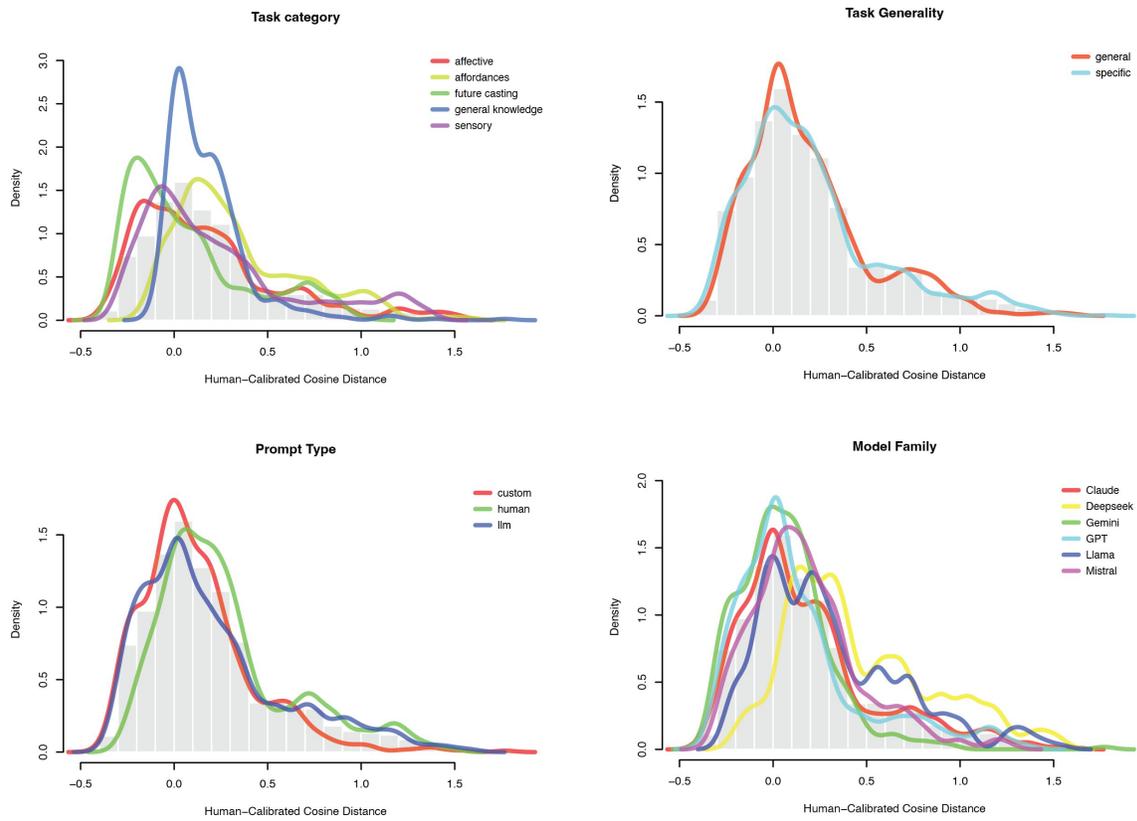

Figure 5: HCD distances among four different groups: Task Category, Task Generality, Prompt Type, and Model Family. Histograms show counts of HCD values, and curves represent kernel density for each group.

| HCD | Navigation Task | Sitting Task |
|---|---|---|
| | 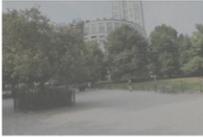 | 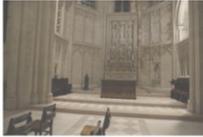 |
| <0 | As you walk along the path, you'll encounter a grassy area surrounded by trees on either side. Ahead, there is a tall, curved building that rises up in the distance. The path continues straight, with benches lining the walkway for you to rest along the way. (Claude 3.0 Haiku, HCD = -.12) | I might choose to sit on one of the wooden chairs nearby or on the benches along the walls. (Magistral Medium, HCD = -.14) |
| 0-.2 | Walk through the gravel path and keep your head up, following the fence on your right. Keep your steps light and even, and move forward slowly. Keep an eye on the people around you to ensure everyone can see you. (Deepseek Janus 1B, HCD = .11) | I would likely choose to sit on the straight back chair on the right side of the image. This chair appears to be designed for comfort and provides a good view of the altar and other religious artifacts. Additionally, it is positioned close to the front of the church, making it accessible for any prayers or services taking place. The other available surfaces, such as the couch in the back left corner, might be more suitable for a longer rest or relaxation after an active period of worship. (Deepseek Janus 1B, .HCD = .02) |
| .2-.4 | As you walk forward, feel the soft gravel underfoot and listen for the sound of rustling leaves from trees on your left. Keep an eye out for a bench to your right, and continue straight until you reach a grassy area, where you'll notice the gentle sounds of people talking and occasional laughter around you. (GPT-4o-mini, HCD =.31) | I might choose to sit on one of the wooden benches in the foreground, as they appear to be available and intended for seating. Another option could be the steps leading up to the altar, although it might be less comfortable due to the hard stone surface. The red cushions on the floor to the right of the altar could also provide a place to sit, likely offering more comfort than the steps. (Pixtral Large, HCD = .26) |
| .4-.6 | As you walk, look for a person sitting on a bench. Approach them slowly and speak to them, introducing yourself and asking for their name. They will then help you navigate around the park and to the building. Keep moving steadily and be mindful of the trees and fence to ensure a safe and pleasant walk. (Deepseek Janus 1B, HCD = .54) | The wooden chairs are clearly the intended and most comfortable option for resting while taking in the beautiful carved stonework and peaceful atmosphere of this cathedral chapel. (Claude Sonnet 4, HCD = .45) |
| .6-.8 | This is an outdoor park with a large, open sandy area in the foreground. There are benches to your right and a fence bordering the sandy area. Behind the fence, there are trees and a few people scattered throughout the park. (Pixtral Large, HCD = .64) | I would avoid sitting on any of the surfaces visible in the image, as this appears to be a religious space, likely a church or cathedral. The ornate altar and architecture suggest this is a sacred space meant for worship, not casual seating. While there are a few chairs visible, they seem to be reserved for clergy or other religious officials. Out of respect for the solemnity of the environment, I would refrain from using any of the available surfaces as casual seating and simply observe the grandeur of the space while standing or slowly walking through it. (Claude 3.0 Haiku, HCD = .64) |
| .8-1 | The building in the background is white and has a lot of windows, so it's easy to spot from a distance. (Magistral Medium, HCD = .85) | Upon entering the scene, I notice that the central focus of the image is the ornate altar, which is adorned with intricate carvings and decorations. It appears to be a significant and sacred space, suggesting that it may be used for religious services or ceremonies. The altar is elevated on a platform with steps leading up to it, implying a focal point for those attending. In the foreground, there are a few chairs arranged around the perimeter of the room. On the left side of the image, there is a plain wooden chair that seems to be available for anyone who wishes to sit there (Deepseek Janus 1B, HCD = .92) |
| >1 | Maybe: (Magistral Medium, HCD = 1.56) | As an AI language and vision assistant, I can assist you in various tasks, but I am not capable of sitting or choosing surfaces to rest. You can describe your choice to me, and I will provide assistance.(Deepseek Janus 1B, HCD = 1.03) |

Figure 6: Representative descriptions for two example images across a spectrum of HCD values for the navigation and sitting tasks.

= 2.81, p = 0.011), indicating that even under baseline conditions, VL outputs were reliably above the human lower-bound anchor.

**Prompting Improved Performance for Non-Affordance Tasks**

We next sought to establish whether certain tasks benefitted from different prompting strategies. While the prompting strategy had no overall effect in general knowledge tasks (both p > 0.37), the prompting strategy differed significantly by task category, as confirmed by estimated marginal means comparisons (Tukey-adjusted) within each task category. In affective tasks, both the custom (GPT-4o–optimized) prompt and VL-generated prompts reduced HCD relative to the human baseline (human–custom=0.23, p<0.001; human-VL=0.14, p=0.0002), and custom prompting also outperformed VL-generated prompts (custom–VL=-0.09, p=0.029). The same pattern held for sensory tasks (human-custom=0.27, p<0.001; human-VL=0.13, p=0.0006; custom-VL=-0.14, p=0.0002). For future-casting tasks, custom prompting reduced HCD relative to both the human baseline (human–custom=0.18, p<0.001) and VL-generated prompts (custom–VL=-0.11, p=0.0043), while the human–VL contrast was not reliable (p=0.09). Critically, prompt differences were not detected for affordances. No other interaction terms were statistically significant (see Table 2). This suggests that while VLM performance can be improved with better prompting for many tasks, performance on affordances was not improved by changing prompting strategy.

Finally, it is worth noting that this model explained moderate variance overall (conditional $R^2$ = 0.26; marginal $R^2$ = 0.11), with additional variation captured by random intercepts for Task and Model. Therefore, we will examine model-level predictors in a second set of analyses.

Table 2: Statistical results on linear mixed effect model predicting HCD from task and prompt conditions. The intercept represents the general knowledge task, the human prompt, and the general tasks.

| Factor | Estimate | Standard Error | t-value | p-value |
| --- | --- | --- | --- | --- |
| Intercept | 0.1700 | 0.063 | 2.811 | 0.0110 |
| Prompt (custom versus human) | -0.0360 | 0.041 | -0.900 | 0.3700 |
| Prompt (VL versus human) | -0.0180 | 0.041 | -0.450 | 0.6500 |
| Task category (affective versus general knowledge) | 0.1100 | 0.066 | 1.590 | 0.1400 |
| Task category (affordances versus general knowledge) | 0.1700 | 0.067 | 2.590 | 0.0220 |
| Task category (future casting versus general knowledge) | -0.0400 | 0.070 | -0.582 | 0.5700 |
| Task category (sensory versus general knowledge) | 0.1600 | 0.070 | 2.470 | 0.0280 |
| Specific versus general | 0.0200 | 0.040 | 0.490 | 0.6300 |
| Custom prompt : affective task | -0.1800 | 0.050 | -3.670 | 0.0002 |
| VL prompt : affective task | -0.1300 | 0.050 | -2.650 | 0.0080 |
| Custom prompt : affordance task | 0.0002 | 0.050 | -0.040 | 0.9700 |
| VL prompt : affordance task | -0.0110 | 0.050 | -0.227 | 0.8200 |
| Custom prompt : future casting task | -0.1300 | 0.050 | -2.660 | 0.0080 |
| VL prompt : future casting task | -0.0600 | 0.050 | -1.270 | 0.2000 |
| Custom prompt : sensory task | -0.2100 | 0.050 | -4.415 | 0.0001 |
| VL prompt : sensory task | -0.1200 | 0.050 | -2.430 | 0.0150 |
| Custom prompt : general/specific | -0.0300 | 0.030 | -1.074 | 0.2830 |
| VL prompt : general/specific | 0.0140 | 0.030 | 0.446 | 0.6560 |

**Newer Models are Improving on Non-Affordance Tasks**

Is VLM performance improving over time, and if so, does this improvement benefit all tasks? We fit a second model with VLM attributes as fixed effects (Release date, operationalized as months since model release (scaled and centered), and Model Family) and their interactions with task structure, while retaining prompt type, and a random intercept for task.

Interestingly, we found no evidence that VL model performance was changing over time ($\beta$ = -0.011, 95% CI [-0.05, 0.02], t(2402) = -0.903, p = 0.37) in the baseline condition. Note that category-specific time trends were evaluated using estimated slopes of HCD as a function of Release within each category. Because Release was coded as months since release, positive slopes indicate worse performance for older models (i.e., improvement for more recent releases). Estimated marginal trends revealed category-specific improvements with model release date (HCD decreased in newer models for affective (slope = 0.039 (95% CI [0.009, 0.07])) and sensory tasks (slope = 0.043 (95% CI [0.01, 0.07])). However, no improvement over time was observed for affordance tasks (slope = 0.02, 95% CI [-0.002, 0.06]), future-casting: slope = 0.028 (95% CI [-0.002, 0.06])), or general knowledge (slope = -0.005, 95% CI [-0.03, 0.02]). Therefore, while VL performance is improving in some tasks, we observed no evidence for improved performance in affordance tasks in newer models, even though there was the most room to improve.

Echoing the results of the first model, we found a significant main effect of the affordance tasks on HCD, even after controlling for prompt strategy, model family, and time since model release ($\beta = 0.19$, 95% CI [0.07, 0.31], $t(9) = 3.08$, $p = 0.011$). By contrast, other task categories did not differ reliably from the general knowledge baseline (see Table 3). As observed in the first model, human-calibrated distance was higher for affordances than the mean of the other four task categories (planned contrast: $\Delta = 0.17$, SE = 0.048, $t(9) = 3.48$, $p = 0.006$). Differences between specific and general tasks were not reliable ($\beta = 0.009$, 95% CI [-0.07, 0.09], $t(9) = 0.233$, $p = 0.82$). Prompt engineering had a large, robust effect: both custom and VL-generated prompts reduced HCD (custom: $\beta = -0.164$, 95% CI [-0.19, -0.13], $t(2402) = -10.6$, $p < 2e-16$; VL: $\beta = -0.072$, 95% CI [-0.10, -0.04], $t(2402) = -4.67$, $p = 3.09e-06$), with the custom prompt yielding the greatest improvement (all pairwise $p < 0.0001$).

**Family Differences in VL Models are Substantial**

How do different VLMs differ from one another? We found substantial differences in HCD across VL model families. Compared to the Claude baseline, we found less human-like responses for Deepseek ($\beta = 0.44$, 95% CI [0.36, 0.52], $t(2402) = 10.77$, $p < 2e-16$), Mistral ($\beta = 0.11$, 95% CI [0.05, 0.18], $t(2402) = 3.68$, $p = 0.0002$), and Llama ($\beta = 0.21$, 95% CI [0.13, 0.29], $t(2402) = 5.31$, $p = 1.2e-07$) models, and higher human-likeness for Gemini ($\beta = -0.12$, 95% CI [-0.16, -0.08], $t(2402) = -6.03$, $p = 1.94e-09$) and GPT ($\beta = -0.058$, 95% CI [-0.10, -0.02], $t(2402) = -3.01$, $p = 0.003$) models. No other interaction terms were statistically significant, see Table 3.

Table 3: Statistical results predicting HCD values as a function of prompt conditions and model properties. The intercept corresponds to the human prompt, the general knowledge tasks, general tasks, centered time since model release = 0.0 (about 14 months), and the Claude model family.

| Factor | Estimate | Standard Error | t-value | p-value |
| --- | --- | --- | --- | --- |
| Intercept | 0.230 | 0.050 | 4.360 | 1.00e-03 |
| Prompt (custom versus human) | -0.160 | 0.015 | -10.590 | 0.00e+00 |
| Prompt (VL versus human) | -0.070 | 0.015 | -4.670 | 3.10e-06 |
| Task category (affective versus general knowledge) | 0.002 | 0.060 | 0.030 | 9.77e-01 |
| Task category (affordances versus general knowledge) | 0.190 | 0.060 | 3.079 | 1.20e-02 |
| Task category (future casting versus general knowledge) | -0.100 | 0.060 | -1.560 | 1.50e-01 |
| Task category (sensory versus general knowledge) | 0.050 | 0.060 | 0.729 | 4.82e-01 |
| Specific versus general | 0.010 | 0.040 | 0.233 | 8.21e-01 |
| Time since model release | -0.010 | 0.020 | -0.650 | 5.17e-01 |
| Open weights (versus closed) | -0.160 | 0.050 | -3.485 | 5.00e-04 |
| Deepseek (versus Claude) | 0.440 | 0.040 | 10.770 | 0.00e+00 |
| Gemini (versus Claude) | -0.120 | 0.020 | -6.030 | 0.00e+00 |
| GPT (versus Claude) | -0.060 | 0.020 | -3.010 | 2.00e-03 |
| Llama (versus Claude) | 0.210 | 0.040 | 5.310 | 1.00e-07 |
| Mistral (versus Claude) | 0.120 | 0.030 | 3.680 | 2.40e-04 |
| Affective task: time since model release | 0.040 | 0.020 | 2.180 | 3.00e-02 |
| Affordance task : time since model release | 0.030 | 0.020 | 1.590 | 1.10e-01 |
| Future casting task: time since model release | 0.030 | 0.020 | 1.620 | 1.10e-01 |
| Sensory task : time since model release | 0.050 | 0.020 | 2.360 | 2.00e-02 |
| General/specific : time since model release | 0.010 | 0.010 | 0.903 | 3.67e-01 |
| Affective task : open weights | 0.005 | 0.040 | 0.112 | 9.11e-01 |
| Affordance task : open weights | -0.070 | 0.040 | -1.590 | 1.10e-01 |
| Future casting task : open weights | -0.020 | 0.040 | -0.397 | 6.91e-01 |
| Sensory task : open weights | 0.030 | 0.040 | 0.583 | 5.60e-01 |
| General/specific : open weights | 0.020 | 0.030 | 0.555 | 5.79e-01 |

## Discussion

We compared human and VL model generated text to a set of previously unreleased real-world photographs across a battery of high-level scene understanding tasks. We found that VLM descriptions were dramatically longer than human descriptions (~4x more words) but less informationally efficient: VLM descriptions exhibited lower lexical diversity across several measures (lower entropy, lower type-token ratio at matched length). Additionally, VLM descriptions had more positive sentiment compared to their human counterparts. A key finding was that affordance tasks yielded the least human-like text from VLMs. Intriguingly, this deficit persisted across prompt engineering conditions, and were not improving in newer VLMs.

Unsurprisingly, we found that, compared to providing VLMs with the same prompt as human observers, custom prompt engineering improved VLM outputs for affective, sensory, and future-casting tasks. This suggests that, for most tasks, VLMs have latent capability that can be elicited by better prompts. Critically, we did not find that custom prompting improved the human likeness of affordance descriptions. Similarly, although newer VLMs improved in their performance on sensory and affective tasks, we observed no significant improvement in affordance or future casting tasks over time. Whatever is driving general VLM improvement (more data, larger models, better training) is not addressing the affordance deficit. It therefore may be the case that scene affordance knowledge needs qualitatively different training from what has been standard.

The low information density of VLM scene descriptions observed here extends previous observations of low information density in large language models (LLMs) more generally (Guo et al., 2023; Shaib, Chakrabarty, Garcia-Olano, & Wallace, 2026). Although LLM-generated text can be rated more highly by human observers than human-generated text in domains such as creative writing (Doshi & Hauser, 2024) or argumentative essays (Herbold, Hautli-Janisz, Heuer, Kikteva, & Trautsch, 2023) that do not require specific factual details, LLM-generated text is generally characterized by internal redundancy and a paucity of specific details (Shaib et al., 2026), colloquially referred to as "slop".

Altogether, the results of Experiment 1 demonstrate that although VLM-generated scene descriptions can be very humanlike in some tasks, a persistent gap exists for affordance tasks. The goal of Experiment 2 is to begin to understand why. We will begin by characterizing the structural and stylistic differences between human- and VLM-generated scene descriptions, and then test six different explanatory hypotheses for these differences.

# Experiment 2: Explaining Performance Patterns Across Tasks

Experiment 1 demonstrated that, across models, prompt strategies, and encoders, VL models struggled to describe scene affordances in a human-like way. The goal of Experiment 2 was to provide insights into why this might be. Figure 7 illustrates our approach. In Section 1, we will characterize the differences between VL- and human-generated text using NLP to assess the differences between human and VLM affordance-based scene descriptions. In Section 2, we will explore six different mechanistic hypotheses to explain why VL models generate descriptions of scene affordances that markedly differ from human observers'.

## Characterizing the differences in affordance descriptions

**Do VL models differ stylistically from humans in affordance tasks?**

We examined whether affordance task descriptions were outliers for VL models in any of our NLP metrics. Any differences here would reflect stylistic differences between VL models and humans that uniquely affected the affordance tasks.

We observed no significant differences in log description length between any of the task groups (all corrected pairwise p>0.25, see Supplemental Item 4). Further, we observed no differences in lexical entropy across task groups (all corrected pairwise p>0.20, see Supplemental Item 5). Similarly, the affordance tasks did not differ significantly from other tasks according to Word2vec cosine similarity (all corrected pairwise p>0.17, see Supplemental Item 6). Finally, we observed no significant sentiment differences between affordance descriptions and the descriptions for other tasks (all corrected pairwise p>0.77, see Supplemental Item 7). Therefore, human-VL affordance description differences were not driven by differences in style that included description length, description complexity, or sentiment, and therefore reflect deeper, more structural differences between VLMs and humans.

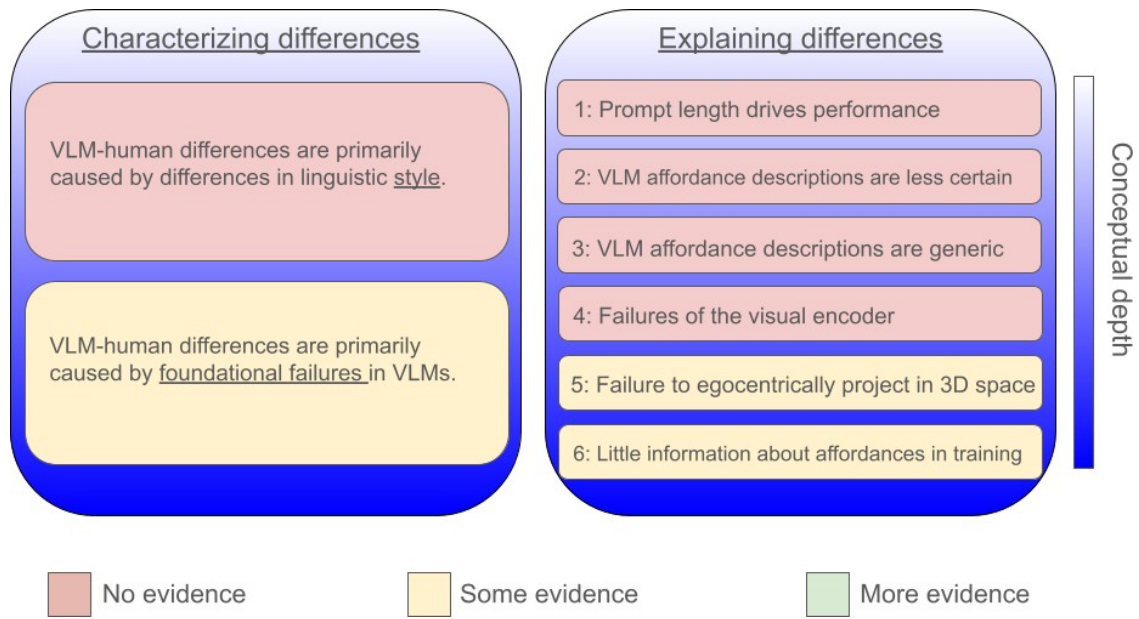

Figure 7: The left side illustrates two basic characterizations of human-VLM differences tested in Section 1 of Experiment 2, while the right side illustrates the six mechanistic hypotheses tested in Section 2 of Experiment 2. For each, boxes are colored by the amount of evidence for each hypothesis and ranked from top to bottom by conceptual depth.

**Are VL affordance differences characterized by catastrophic failures?**

The NLP analyses ruled out purely stylistic reasons for the affordance performance gap. We next sought to assess the extent to which poor performance reflected a genuine failure to understand the affordance tasks. If VL models' failures with the affordance tasks represent a true lack of capability in these tasks, then we should observe a higher rate of catastrophic failures to replicate human descriptions (i.e., HCD values greater than 1), resulting from hallucinations, task refusal, or other serious deficits.

Human-calibrated distances could be greater than one, reflecting cases where the average VL model response for a given image was more different from the human response than human responses for an entirely different image was to the human response. If VL models' poorer performance on affordance tasks is due to a foundational lack of task understanding, then we would expect a higher proportion of HCD values above one for affordances than for other tasks.

To test this, we examined the proportion of HCD values exceeding 1 across Task Groups, Prompts, Model Families, and Time since Model Release. This model had a conditional $R^2$ of 0.62, with a marginal $R^2$ of 0.53 attributable to fixed effects alone. The model's intercept, corresponding to Task Group = general knowledge, Prompt = human, Time since Model Release = 0 and Model Family = Claude, is at -4.42 (95% CI [-5.98, -2.87], $p < .001$). This corresponds to a baseline catastrophic failure rate of 1.2%.

We found that catastrophic failures were largely driven by the affordance and sensory tasks. While all Task Groups showed low proportions of greater-than-one HCDs, general knowledge, affective, and future casting had the lowest proportions (1.2%, 4.9%, and 0.2%, respectively). Affordance and sensory tasks accounted for 6.2% and 8.6%, respectively. Within Task Groups, relative to the general knowledge baseline task, we found that affordances and sensory tasks showed significantly higher catastrophic failure rates (affordances: $\beta = 2.09$, 95% CI [0.28, 3.90], $p = 0.024$, sensory: $\beta = 2.47$, 95% CI[0.67, 4.26], $p = 0.007$). By contrast, affective and future casting did not differ significantly from the general knowledge baseline (affective: $\beta = 1.34$, 95% CI [-0.47, 3.15], $p = 0.148$, future casting: $\beta = -1.64$, 95% CI [-4.28, 1.00], $p = 0.222$). Follow-up pairwise tests using Holm correction revealed that affordances had higher proportions than future casting (affordances - future casting = 3.73, $p = 0.031$) and sensory had higher proportions than future casting (future casting - sensory= -4.11, $p = 0.012$). This means that affordances and sensory tasks were more likely to output responses that were less similar to human responses than an entirely different image, likely reflecting a fundamental issue in the VL models' understanding of the tasks. Overall, although affordance tasks had a higher rate of catastrophic failures than some tasks, sensory tasks were also difficult by this metric. Therefore, the deficits in VL models' affordance descriptions are likely to have multiple causes.

Catastrophic failures also varied between prompt conditions and model families. Within prompt conditions, we found that, compared to the human prompt, there were significantly higher greater-than-one HCD proportions for the custom prompt ($\beta$ = -1.87, 95% CI [-2.63, -1.10], p = 1.61e-06). The VL-generated prompt was not significantly different from the human prompt. Within model families, both Deepseek and Gemini differed significantly from the Claude baseline. Deepseek had significantly higher greater-than-one HCD proportions ($\beta$ = 1.33, 95% CI [0.63, 2.04], p = 0.0002) and Gemini had significantly lower greater-than-one HCD proportions ($\beta$ = -2.89, 95% CI [-4.94, -0.84], p = 0.006). There was no main effect of Time Since Model Release. Therefore, while prompting can prevent catastrophic failures to a certain extent, model differences are widespread and there is no strong evidence that models are improving in this metric over time.

## Mechanisms of VL and human differences in affordances

The previous analyses indicated that the affordance-based scene description differences observed in Experiment 1 were driven more by structural failures than superficial stylistic differences. The goal of this section is to explore several mechanistic hypotheses that could explain these differences. Here, we roughly order the hypotheses by the depth of representational failure, with early hypotheses reflecting engineering constraints that are likely to be resolved with increasing resources, and later hypotheses reflecting a foundational lack of affordance-based scene understanding in VL models.

### Hypothesis 1: Performance across tasks could be attributed to prompt length

Some models (especially in the Deepseek family) had very small context windows (see Supplemental Item 1). If a prompt's length approaches the VLM's context window limit, then fewer tokens can be used in the scene description, and this could lead to a less human-like performance. If this is the case, we would expect a negative correlation between prompt length and a VL model's HCD. More specifically, we hypothesized that if prompt length is driving the previously-observed task differences, we would observe longer prompts for the affordance tasks. In order to test this hypothesis, we compared the length of each prompt (in characters) across task groups. Prompt length varied from an average of 924 characters for affective to 1124 for future casting. However, compared with a baseline for general knowledge tasks, we observed no reliable differences across task groups (all p > 0.4; see Supplemental Item 8 for all statistics). Therefore, the task-related differences we observed cannot be attributed to prompt length.

**Hypothesis 2: VL models express more uncertainty for affordance tasks**

Alternatively, VL models may exhibit greater expressions of uncertainty in affordance tasks than in other task groups. To test this hypothesis, we created a lexicon of 32 words associated with hedging behavior (e.g., "maybe", "possibly", "seems", see Supplemental Item 9 for complete list), and compared the proportion of descriptions with hedging language across tasks. To model hedging as a continuous outcome while respecting its bounded range [0,1], we applied a logit transformation to the hedging proportion. Because the logit is undefined at 0 and 1, we first winsorized the proportions to lie within [$\epsilon$, 1-$\epsilon$] using a small constant $\epsilon$ = 10e-6. Using a linear mixed-effects model with random intercepts for Task and Image, we found that human descriptions contained significantly less hedging than VL model outputs ($\beta$ = -0.464 on the logit scale, SE = 0.136, t(32945) = -3.43, p = 0.00061), corresponding to ~37% lower hedging odds. This suggests that VL models' descriptions can be characterized by a good deal of expressive uncertainty overall. To test whether affordance tasks disproportionately caused this hedging behavior, we fit a linear mixed-effects model predicting logit-transformed hedging rate from Task Group, with random intercepts for Task and Model. Estimated marginal means and Tukey-adjusted pairwise comparisons showed that general knowledge descriptions contained significantly less hedging than affective ($\Delta$logit = 4.50, p = .0126), future-casting ($\Delta$logit = 5.30, p = .0016), and sensory tasks ($\Delta$logit = 4.54, p = .0112), consistent with these task groups eliciting more uncertainty/qualification language overall. In contrast, affordances did not differ reliably from general knowledge ($\Delta$logit = 2.08, p = .578) and did not differ from the other task groups after correction (all p > .15). Thus, while hedging varies substantially by task demands, affordance tasks are not uniquely characterized by elevated hedging, suggesting that VL models' differences in affordance tasks was not driven by a higher expression of uncertainty.

**Hypothesis 3: VL models' affordance knowledge is shallow and associative**

If VL models' poorer performance on affordance tasks is due to shallow, associative affordance knowledge, then one might expect that VL descriptions contain more stereotyped and generic responses for affordance tasks compared to others. In this case, we would expect a higher proportion of negative HCD values than in other tasks. Negative HCD values occur in cases when human responses were more diverse than those produced by the VL models. We examined the proportion of negative HCD values across Task Groups, Prompts, Model Families, and Time since Model Release, including Task as a random effect. This model had a conditional $R^2$ of 0.39, with a marginal $R^2$ of 0.35 attributable to fixed effects alone. The model's intercept, corresponding to Task Group = general knowledge, Prompt = human, Time since Model Release = 0, and Model Family = Claude, is at -2.58 (95% CI [-3.20, -1.97], p < .001). This corresponds to an overall negative HCD proportion of 7.5%.

Interestingly, we found general knowledge and affordances had a relatively low percentage of negative HCDs (14.0% and 13.8%, respectively). By contrast, affective (40.7%), future casting (54.7%, and sensory (37.4%) elicited more negative HCDs. Within Task Groups, relative to the general knowledge baseline task, we found that affordances did not differ significantly in the proportion of negative HCDs ($\beta$ = -0.001, 95% CI [-0.63, 0.63], p = 0.996). By contrast, affective ($\beta$ = 1.67, 95% CI [1.07, 2.27], p = 5.68e-08), future casting ($\beta$ = 2.36, 95% CI [1.75, 2.96], p = 2.06e-14), and sensory ($\beta$ = 1.50, 95% CI [0.89, 2.10], p = 1.15e-06 tasks had higher proportions of negative HCDs relative to general knowledge. Follow-up pairwise tests using Holm correction revealed that these affordances also have significantly lower proportions than affective (affective - affordances = 1.67, p < 0.0001), future casting (affordances - future casting = -2.36, p < 0.0001), and sensory (affordances - sensory = -1.50, p = <0.0001). This indicates that affordance descriptions from VL models were more lexically diverse than those for affective, future-casting, and sensory tasks, but not different from general knowledge tasks.

We also found diversity in the negative HCD rate across prompt conditions and model families. Within the prompt conditions, we found that, compared to the human prompt, there were significantly higher negative HCD proportions for both the custom prompt ($\beta$ = 1.09, 95% CI [0.84, 1.34], p <2e-16) and the VL-generated prompt ($\beta$ = 0.90, 95% CI [0.64, 1.15], p = 3.53e-12). This indicates that there was less diversity in responses to these kinds of prompts than to human prompts. Within Model Families, compared with the Claude family, the Deepseek models had a significantly lower proportion of negative HCD descriptions ($\beta$ = -2.31, 95% CI [-3.12, -1.51], p = 1.69e-08). No other reliable differences were observed across model families, see Table 4. Within Time Since Model Release, there is no overall main effect of Time Since Model Release, but we did observe significance for time since release within the LLM prompt ($\beta$ = -0.38, 95% CI [-0.63, -0.14], p = 0.002) This showed that newer models have higher proportions of negative HCDs for the LLM prompt relative to the human prompt. Overall, these results do not suggest that the failure of VL models on affordance tasks was driven by the creation of overly generic descriptions.

Table 4: Statistical results predicting negative HCD values as a function of prompt conditions and model properties. The intercept corresponds to the human prompt, the general knowledge tasks, general tasks, centered time since model release = 0.0 (about 14 months), and the Claude model family.

| Factor | Estimate | Standard Error | t-value | p-value |
| --- | --- | --- | --- | --- |
| Intercept | -2.580 | 0.32 | -8.190 | 0.00e+00 |
| Affective task (versus general task) | 1.670 | 0.31 | 5.430 | 1.00e-07 |
| Affordance tasks (versus general knowledge) | -0.001 | 0.32 | -0.005 | 9.96e-01 |
| Future casting task (versus general knowledge) | 2.360 | 0.31 | 7.640 | 0.00e+00 |
| Sensory tasks (versus general knowledge) | 1.500 | 0.31 | 4.860 | 1.20e-06 |
| Custom prompt (versus human) | 1.090 | 0.13 | 8.540 | 0.00e+00 |
| VL prompt (versus human) | 0.900 | 0.13 | 6.960 | 0.00e+00 |
| Time since model release | 0.010 | 0.19 | 0.070 | 9.50e-01 |
| Deepseek family (versus Claude) | -2.310 | 0.41 | -5.640 | 0.00e+00 |
| Gemini family (versus Claude) | 0.470 | 0.31 | 1.530 | 1.30e-01 |
| GPT family (versus Claude) | 0.330 | 0.29 | 1.130 | 2.60e-01 |
| Llama family (versus Claude) | -0.580 | 0.39 | -1.490 | 1.40e-01 |
| Mistral family (versus Claude) | -0.500 | 0.32 | -1.550 | 1.20e-01 |
| Affective : time since model release | -0.008 | 0.17 | -0.049 | 9.60e-01 |
| Affordances : time since model release | -0.080 | 0.19 | -0.420 | 6.70e-01 |
| Future casting : time since model release | 0.140 | 0.17 | 0.820 | 4.10e-01 |
| Sensory : time since model release | -0.200 | 0.17 | -1.190 | 2.40e-01 |
| Custom prompt : time since model release | 0.030 | 0.12 | 0.240 | 8.10e-01 |
| VL prompt : time since model release | -0.380 | 0.13 | -3.040 | 2.00e-03 |

**Hypothesis 4: VL model failures for affordances reflect failures of the visual encoder**

To successfully describe scene affordances such as its possibilities for navigation and sitting, a model must represent the scene's 3D spatial layout from the 2D photograph, and link that layout to the capabilities of a human-scaled agent. Recent work has shown that VL models struggle with basic visual tasks (Tangtartharakul & Storrs, 2026; Tong et al., 2024) such as assessing numerosity or determining if two lines are parallel. If VL models do not receive sufficient spatial information about a scene from their visual encoder, they may fail to represent the three-dimensional scene from the two-dimensional picture, and thus disproportionately struggle to describe scene affordances. To test this hypothesis, we replaced the images in the prompts with spatially explicit scene descriptions.

One author (MRG) created general descriptions for each scene that explicitly provided absolute and relative spatial information. For example, for the image on the right-hand side of Figure 6, "You are near the front of an elaborate church or cathedral. There are chairs immediately in front of you and to the left. The altar is in front of you, about 25 meters away (and up two shallow stairs). There is seating on either side of the altar about 5-10 meters away. Beyond the altar is a pipe organ. The walls on the far side have elaborate carvings and there are stained glass windows near the ceiling. The floor is marble tile." These descriptions were meant to be more detailed than the general scene descriptions obtained online (typical example of the same image: "inside a large church, some wooden chairs at the front with a pulpit further back"). We provided these descriptions to six VL models (the most recently released model within each family), along with human-prompt instructions for the navigation and sitting tasks. Human-calibrated cosine distances were computed as in Experiment 1.

If HCD values improve when replacing the image with a spatially-explicit text description, this would indicate that the visual module contains insufficient spatial information, but that affordance information could be recovered when this information is provided via text. However, we did not find evidence of a benefit: whereas the original average HCD was 0.16 across the two tasks, the HCDs for the spatially-explicit text increased to 0.22, indicating that the photograph provided helpful information to the VLMs. A mixed-effects model predicting HCD from Condition (text versus image), Task (navigation versus sitting), and their interaction, with random intercepts for Image and Model, revealed no main effect of Condition ($\beta = 0.001$, 95% CI [-0.007, 0.009], $p = 0.77$) but a strong Condition x Task interaction ($\beta = -0.13$, 95% CI [-0.14, -0.12], $p < 2e-16$). For navigation descriptions, the intervention had essentially no effect on HCD. For sitting descriptions, however, the intervention substantially increased HCD relative to the original prompt, indicating less human-like responses.

To ensure that the spatially explicit scene descriptions contained sufficient information to replace the original images for the affordance tasks, we conducted a validation study with human participants. 800 participants were recruited on Cloud ResearchConnect to provide descriptions for the sitting and navigation tasks based on the spatially explicit scene descriptions instead of the images. 445 participants identified as women, 329 as men, 18 as non-binary, 1 as agender, 4 as members of non-listed gender minorities, and 3 as not disclosing their gender. Participants ranged in age from 18 to 78 years (mean=37.6 years, median = 35 years, sd = 12.33). Other than replacing the images with scene descriptions, the participants' instructions remained unchanged. Each individual contributed between 1 and 40 descriptions to provide a total of 10 human descriptions per image per task. If human observers could provide equally good responses with either the image or the longer text, we would expect HCD values of 0.0. We found that the average HCD between the text descriptions and the original image descriptions for humans was 0.11 (sitting HCD = 0.04, navigation HCD = 0.18). Therefore, although human observers benefitted from the image, the fact that HCDs in the text condition were lower than the VL models in the text condition suggests that the spatially explicit scene descriptions contain sufficient information for humans to recover affordance information. Given that the spatially explicit descriptions were sufficient to recover affordance information but the VL models were unable to make use of it, we conclude that their affordance description differences were not due to a failure of the models' visual encoders.

### Hypothesis 5: VL models fail to describe affordances because they do not automatically project themselves into the three-dimensional structure of a scene from a photograph

Affordance understanding is an intrinsically egocentric act. Therefore, if a VL model does not automatically project itself into the three-dimensional structure of a scene, it may not be able to express affordance-related information. To test this, we provided the same six models with amended human prompts for affordance tasks that explicitly provided viewpoint information.

We amended the prompts to include information on the viewer's position within the scene. For example, the amended navigation prompt read "*You are standing where the camera is, facing into the scene.* Imagine that you are telling a blind person next to you how to walk through this scene. Write a short (1-3 sentence) description that could instruct such a person." (Italicized text shows addition from original prompt). We provided this prompt to the same six models from Hypothesis 4, and obtained descriptions, computing human-calibrated cosine distances as in Experiment 1.

Interestingly, we found that providing viewpoint information provided an overall benefit to VL models. A mixed-effects model predicting human-calibrated cosine distance from Condition (original versus amended), Task (navigation versus sitting), and their interaction, with random intercepts for image and model, revealed a significant main effect of Condition ($\beta = 0.007$, 95% CI [3.09e-04, 0.01], $p = 0.04$), indicating significant improvement in HCD with the amended prompt. Further, we found that the sitting task led to consistently worse performance compared to navigation ($\beta = 0.08$, 95% CI [0.07, 0.08], $p < 2e-16$), and a significant interaction between condition and task, indicating that the amended prompt improved performance in the sitting task more than navigation ($\beta = -0.14$, 95% CI [-0.15, -0.13], $p < 2e-16$). In a separate analysis, we examined the rates of generic responses (negative HCDs) and catastrophic failures (HCDs>1) as a function of prompt condition. We found that the original prompt condition increased the rate of catastrophic failures ($β = 0.23$, 95% CI [0.04, 0.42], p = 0.017), but had no effect on the rate of negative HCD responses ($β = 0.01$, 95% CI [-0.05, 0.07], p = 0.74), compared to the amended prompt. Therefore, explicitly specifying viewpoint information can aid human-VLM alignment on affordance tasks, but its benefit is highly task-specific.

**Hypothesis 6: VL models' training data contains sparse information about scene affordances**

Humans are highly sensitive to scene affordances (Gibson, 1986; Greene et al., 2016; Greene & Hansen, 2025), and Gricean maxims dictate that humans do not explain things that are self-evident to other humans (Grice, 1991). Therefore, image caption datasets used to train VL models may have comparatively less information about scene affordances. The lack of scaffolding of affordance information in language could cause lower performance in the affordance tasks.

Although training data for many closed weight VLMs has not been specified, image captions provide one type of information that is likely part of all VLM training regimes. We examined two image captioning datasets to test this hypothesis. The Conceptual 12M dataset (Changpinyo, Sharma, Ding, & Soricut, 2021) is a set of 12 million image-text pairs sourced from various internet sources, and Redcaps (Desai, Kaul, Aysola, & Johnson, 2021) is a set of an additional 12 million image-text pairs sourced from Reddit. Our approach was to create three lexicons containing words that could be used in affordance descriptions: permission/constraint (e.g., "allow", "can", "caution", etc.), instrumentality (e.g., "use", "tool", "enable", etc.), and accessibility (e.g., "narrow", "path", "open", etc.), for a total of 150 affordance-related terms. As a control, we created two lexicons of affective concepts (20 positive, 20 negative). We reasoned that, because affective tasks were comparatively easier for VL models, a lower rate of affordance terms compared to affective terms might have resulted from statistics of their training sets. To find a set of 40 affordance terms with similar word frequency to the affective term list, we used a quartile matching procedure. We first estimated the word frequency of each affect term using the Brown corpus (Kučera & Francis, 1967) as a reference corpus. We then computed empirical quantiles from this frequency distribution. For each target quantile, q, we computed the corresponding frequency value, f, from the affective lexicon and selected words from the affordance candidate pool whose frequency was closest to f, enforcing selection without replacement to ensure that each candidate word could be chosen at most once. This yielded a set of 40 affordance terms whose frequency distribution closely tracked that of the 40 affect terms. The full set of words in each lexicon can be found in Supplemental Item 10.

In both image caption datasets, matched terms were slightly more likely to belong to the affective lexicon than the affordance (CC12M: 52% affective; RedCaps: 56% affective), despite being matched for baseline word frequency. We compared this rate to the affordance-heavy descriptions from WikiHow (Koupaee & Wang, 2018), finding that only 36% of matched terms were affective, indicating that our affordance lexicon does indeed predominate when affordances are being conveyed. While this analysis demonstrates that there might be slightly less affordance information in image captioning sets, this modest imbalance alone is unlikely to explain VLMs' pronounced affordance deficits.

Affordances are complex relationships between objects and possible actions, and it is difficult to identify them using simple distributional semantics. To test whether image caption datasets contain affordance vocabulary in relational (agent-action-object) constructions, we analyzed the syntactic structure of affordance term usage across CC12M and RedCaps datasets. We used the same affordance and affective lexicons described above and extracted all sentences containing at least one term from either lexicon. For each term occurrence, we used the dependency parser from the spaCy library (en_core_web_sm) to extract syntactic features indicative of relational affordance language: (1) whether the term functioned as a verb versus a noun or adjective, (2) whether the sentence contained second-person pronouns ("you," "your"), indicating an addressed experiencer, (3) whether the term appeared in a modal construction ("can sit," "should use"), encoding possibility or permission, (4) whether the term governed a spatial prepositional complement ("walk through the door"), (5) whether the sentence was imperative (verb-initial with no explicit subject), and (6) whether the term appeared in a purpose clause ("to open," "for walking").

As shown in Table 5, the WikiHow dataset generally contained significantly more of the six affordance construction features than did either of the image caption datasets.

Table 5: Chi square test differences between affordance constructions in image caption datasets versus WikiHow. Positive percent differences indicate more of a feature in WikiHow than a caption set. *** indicates adjusted p«0.0001.

| Feature | Comparison | Percent Difference | Chi Sq | p-value | Cramer's V |
| --- | --- | --- | --- | --- | --- |
| As verb | CC12M vs WikiHow | 11.7 | 19002.32 | *** | 0.11 |
| As verb | RedCaps vs WikiHow | -6.3 | 4593.15 | *** | 0.06 |
| Imperative | CC12M vs WikiHow | 12.9 | 48536.60 | *** | 0.18 |
| Imperative | RedCaps vs WikiHow | 1.1 | 203.90 | *** | 0.01 |
| Modal | CC12M vs WikiHow | 3.3 | 18735.20 | *** | 0.11 |
| Modal | RedCaps vs WikiHow | 2.8 | 8537.90 | *** | 0.09 |
| Purpose | CC12M vs WikiHow | 5.8 | 19301.70 | *** | 0.11 |
| Purpose | RedCaps vs WikiHow | 4.2 | 6876.60 | *** | 0.08 |
| Second person | CC12M vs WikiHow | 90.4 | 1174788.90 | *** | 0.88 |
| Second person | RedCaps vs WikiHow | 93.2 | 989795.10 | *** | 0.93 |
| Spatial | CC12M vs WikiHow | -3.5 | 2351.90 | *** | 0.04 |
| Spatial | RedCaps vs WikiHow | -9.1 | 12240.80 | *** | 0.10 |

Therefore, although there is only a slight tendency towards lower affordance use in image caption datasets overall, the majority of these terms are not used in ways that specifically instruct about affordances. Chi-squared tests comparing relational feature rates between corpora revealed that second-person address was the primary distinguishing feature (CC12M vs. WikiHow: $\chi^2$ = 1,174,789, Cramér's V = 0.88; RedCaps vs. WikiHow: $\chi^2$ = 989,796, V = 0.93; both p < .001). All other features showed small effect sizes (V < 0.18). Notably, spatial prepositional complements were actually more frequent in caption corpora than in WikiHow (CC12M: 24.6%, RedCaps: 30.2%, WikiHow: 21.1%), confirming that image captions encode spatial relations (e.g., "path through the forest") but without linking them to an addressed agent. Therefore, if affordance performance was inherited from sparser affordance-based information from the training regime, it is likely that it emerges from second-person address. In other words, if a VLM is able to model the role of an experiencer who will act on affordances, it may be able to more faithfully describe affordances. Future work should examine whether providing this information in VLM fine tuning improves affordance descriptions.

## Discussion

In Experiment 2, we characterized the features of VLM affordance descriptions that distinguished them from human responses, and analyzed patterns of success and failure to understand the sources of this deficit. Across our explorations, we found little evidence that the affordance deficit was shallow or stylistic. Instead, we found evidence that the deficits were more structural and semantic.

We found that our characterizations and explanations that had less conceptual depth did not have evidence: VL model descriptions for the affordance task group did not differ in linguistic style from other task groups, the prompt lengths for affordance tasks were not significantly different than other tasks, the VL models did not express more uncertainty for affordance tasks, the descriptions were not more stereotyped (i.e, no more negative HCDs), and the failures did not seem to be due to failures in the visual encoding model since removing the need for the visual encoder by swapping out images for spatial descriptions did not improve results.

By contrast, our results pointed to alternatives that suggested that affordance deficits were deeper and more structural. We found that affordance tasks differed reliably in terms of the rate of catastrophic failures from most other tasks: VLMs for affordances had higher rates of HCDs that were greater than 1, as did sensory tasks. This indicates that VLMs will frequently provide text that is more different from human descriptions than a description of an entirely different image. In other words, these failures reflect fundamental model misunderstandings in these tasks.

We tested two interventions to see if they improved the VLMs' performance for the affordance tasks. We found that adding language to the prompt about the viewer's position within the scene helped some models in some tasks produce output that is more similar to human output. We also found that replacing the images with spatial descriptions did not improve VLM output similarity to humans. These results show that while spatial information alone is not enough for VLMs to mirror human responses, including information that primes the VLM that there is an agent and the agent has a size and a perspective can yield better results. This ties to our finding that while affordance language is present in image captioning corpora, the majority of this language does not include affordance information in relation to an agent. One remaining possibility is that VLMs fail at affordances because they do not have a natural understanding of the physicality of an agent like humans do which is necessary for completing these affordance tasks that are about what an agent can do in a space.

None of these characterizations or hypotheses on their own fully characterize and explain the difference between affordance tasks and other tasks that leads to less human-like affordance descriptions. However, taken together, our findings paint a picture that shows that VLMs lack a deep understanding of what it's like to be an embodied agent, which may be fundamental to performing certain tasks, such as affordance tasks.

# General Discussion

Human scene understanding is deeply linked not only to purely visual processing, but also to a rich, embodied experience of other sensory content, affordances and motor planning, predictions of upcoming events, and our affective reactions to the environment. What information is sufficient to learn this rich set of associations? In this work, we used vision-language models (VLMs) to examine the extent to which scene understanding can be achieved through the joint learning of two-dimensional pictures and text descriptions. In two experiments, we found that performance differed markedly across task types, implying that scene understanding is not unitary. Some information, especially general knowledge, emotional reasoning, and future prediction, appear to be learnable from image-language statistics. However, VLMs persistently struggle to describe scene affordances in a human-like manner. Further, affordance tasks remain harder across prompting conditions, and are not well-explained by shallow, stylistic differences, including the expression of uncertainty or the genericness of the responses. Further, VLM performance on affordance tasks is not improving in newer models. This work evokes previous results with large language models (LLMs), who show deficits in affordance reasoning in text inputs (Jones et al., 2022).

While this work demonstrates that the performance gap for affordance tasks is unlikely to have a single cause, the explorations of the phenomenon in Experiment 2 collectively point to VLMs' reduced ability to project themselves as a human-like agent in a three-dimensional scene from the two-dimensional picture itself. We do not believe that this is a simple failure of the models' visual encoders: when images were replaced with spatially-explicit text that was sufficient for human observers to generate affordances, VLMs were not able to generate better affordance descriptions. Rather, we found that affordance performance improved in many cases when we provided information that explicitly located the viewer at a camera position within a three-dimensional scene. Taken together, these two results imply that merely handing models more spatial text is not enough, but explicitly specifying the viewer's position can help. However, this level of improvement was task- and model-specific, and thus did not fully explain the performance gap. Finally, we found that affordance-relevant language is comparatively sparse in image caption data used for training data, particularly in agent-addressed, action-guiding constructions.

This pattern of results can be explained with two complementary theoretical frameworks. The first, the Gricean hypothesis, posits that foundation models, including VLMs and LLMs should exhibit difficulties precisely in the tasks that are easiest for human observers. Specifically, humans do not describe things that are self-evident to other humans (Grice, 1991). Because human observers are highly sensitive to scene affordances (Gibson, 1986; Greene et al., 2016; Greene & Hansen, 2025), it might be the case that human-generated language contains sparse and compressed affordance information that does not support model learning about affordances. We found that affordance-based information was comparatively sparse in image-captioning datasets, particularly constructions where agents are addressed in the second person. However, given the correlational nature of these results, and their comparatively modest effect sizes, it's unlikely that Gricean language compression is the sole cause of the affordance performance gap.

The second, the embodiment hypothesis, holds that affordances cannot be learned from distributional semantics or flat photographs — they must be grounded in three-dimensional experience. This follows Gibson's original claim that affordances are directly perceived through invariant features arising from bodily movement through the environment (Gibson, 1986). This implies that no two-dimensional proxy can fully substitute for embodied, three-dimensional experience. While the strongest versions of these claims remain controversial, empirical evidence in both human and machine vision supports them. First, compared to two-dimensional images, real objects elicit very different neural responses (Snow & Culham, 2021), and attentional engagement (Gomez, Skiba, & Snow, 2018). Second, object learning in typically-developing infants (Soska et al., 2010) as well as children whose vision has been surgically restored in late childhood (Ostrovsky et al., 2009) is linked to viewing objects in different positions via self motion. Finally, some in computer vision have argued that meaning is a result of grounding symbolic representations via multimodal perceptual experiences of concepts in the world (Bisk et al., 2020), and agentive exploratory experience seems beneficial to VLM learning (Zhang et al., 2026).

In these experiments, VLMs showed underperformance with sensory as well as affordance tasks. Previous work has shown that LLMs use sensory language in a manner that differs from human observers (Hicke, Hamilton, & Mimno, 2025), although models vary in how they differ from humans. However, LLM and VLM performance across sensory tasks is not uniform. While models show strong human alignment in basic sensory features, including color, pitch, and loudness (Kawakita, Zeleznikow-Johnston, Tsuchiya, & Oizumi, 2024; Marjieh et al., 2024), other sensory features, including taste, musical timbre, smell, and tactile features show low human alignment (Marjieh et al., 2024; Zhong, Zhou, Dawes, Brianz, & Obrist, 2024). Interestingly, VLM alignment is not stronger than LLM alignment, even in visual tasks (Marjieh et al., 2024), suggesting that most sensory knowledge in foundation models is gained from text. Future work can examine the extent to which certain types of sensory information are present in training corpora to disentangle the extent to which non-alignment is due to sparse training information, as would be predicted by the Gricean hypothesis, versus the need to train with information beyond text, such as embodied world experience.

Our results also provide insights into applied aspects of VLM use in psychology. For example, our results demonstrate that VLMs diverge markedly from human observers in terms of writing style. Specifically, VLMs generate far more text, but with less information density compared to human observers, consistent with the characterization of generative AI-derived content as "slop" (Shaib et al., 2026). Critically, these stylistic differences are not task-dependent: explorations in Experiment 2 found that stylistic differences could not explain the performance patterns across tasks. These results also highlight the need to show human-relative semantic alignment, such as we did through the human-calibrated cosine distance metric, to disentangle stylistic differences from semantic ones. We also observed some illuminating interactions between semantics and style across tasks. For example, VLMs exhibited lower rates of hedging language for both general knowledge and affordance tasks, likely reflecting the fact that these tasks require more objective information than more speculative tasks, such as those in the future casting or affective groups. However, the VLMs' performance in general knowledge tasks exceeds their performance in the affordance tasks, making these descriptions uniquely read as "confidently wrong". This pattern of results echoes the charge that generative AI produces "bullshit" in the sense of philosopher Harry Frankfurt — highly fluent and confident prose that is utterly unconcerned with its own veracity (Gorrieri, 2024; Hicks, Humphries, & Slater, 2024). It may also be the case that the relative success of VLMs in more subjective tasks can be attributed to this proclivity, or reflective of the fact that human observers and VLMs speculate in similar ways. In this way, our results concur with other recent work that demonstrates that VLMs can emulate human-like visual affective reasoning (Conwell, Graham, Boccagno, & Vessel, 2025).

These results also have implications for psychologists and cognitive neuroscientists who have begun to use VLMs as a research tool. VL models have been used to annotate ground truth categories (Rong et al., 2025), emotional features such as facial expressions (Niu et al., 2025), and semantic descriptions (Murlidaran & Eckstein, 2025). Quality model-generated content could help scale cognitive research, but it is necessary to assess the limitations of these models as well. Our data suggest that some tasks, such as general knowledge tasks, can be done very well by VLMs, although they fall short of fully human-interchangeable by our HCD metrics. However, for researchers interested in affordance and sensory tasks, model-generated content should not substitute for human descriptions.

Although this work has analyzed human and VLM-generated descriptions on the order of hundreds of millions of words, there are critical limitations that should be considered. Firstly, our experiment set was limited to 40 real-world images. Although they were selected to represent a diverse range of places, the small set may not generalize to larger stimulus sets. Further, resource and time constraints required that some of the follow-up interventions from Experiment 2 were done only on a subset of models, limiting inferences that could be made about specific architectures or training regimes playing a causal role in performance patterns. Our design aimed to examine patterns of performance across a broad range of high-level scene tasks. Given the marked deficits for affordances, as well as the differences observed between affordance tasks in Experiment 2, it would be illuminating to examine a more diverse range of affordance tasks in future work. Finally, this work examined human-VLM alignment through assessing the semantic similarity in their generated descriptions. Future work can also examine the internal model states to assess the extent to which affordance information is represented, but not expressed.

Across 18 VLM models from six different model families, and across a broad range of 15 high-level scene understanding tasks, we observed that although VLMs can generate human-like scene descriptions in many cases, persistent gaps exist for affordance and sensory tasks that are structural and semantic rather than stylistic. The affordance gap is particularly telling: it resists prompting interventions, does not improve with model scale, and cannot be attributed to training data sparsity alone. Taken together, these findings suggest that the path to human-like scene understanding may require more than scaling what already exists. To understand an environment's action possibilities may require the kind of agent-centered, three-dimensional experience that no photograph and no caption can fully encode. Whether that gap is bridgeable through simulation, embodied robotics, or some representation we have not yet imagined remains an open question. But the present results suggest it is a real gap, and one worth taking seriously.

## Author Contributions

GR: conceptualization, data curation, formal analysis, investigation, methodology, project administration, visualization, writing: original draft, writing: review and editing. SS: conceptualization, data curation, formal analysis, investigation, writing: original draft, writing: review and editing. BCH: conceptualization, funding acquisition, methodology, writing: review and editing. MRG: conceptualization, data curation, formal analysis, funding acquisition, investigation, methodology, software, supervision, visualization, writing: original draft, writing: review and editing.

## Ethics Statement

## Acknowledgments

This work was funded by NSF-2522311/2522312 to MRG and BCH. The authors wish to thank Pei-Ling Yang, Edward Vessel, and Uri Hasson for helpful conversations about the project.

# Data Availability

All data, stimuli, and source code for this project are available at the Open Science Framework (https://osf.io/87kxh/).